\documentclass{article}
\usepackage{ijcai24}
\usepackage{ijcai24-manipulating-embeddings-stable-diffusion-frame}

\graphicspath{{./ijcai24-manipulating-embeddings-stable-diffusion-figures/}}

\begin{document}
\title{Manipulating Embeddings of Stable Diffusion Prompts}

\author{
Niklas Deckers$^{1,2}$
\and
Julia Peters$^{1,2}$\And
Martin Potthast$^{2,3,4}$\\
\affiliations
$^1$Leipzig University\\
$^2$ScaDS.AI\\
$^3$University of Kassel\\
$^4$hessian.AI\\
}

\maketitle

\begin{abstract}
Prompt engineering is still the primary way for users of generative text-to-image models to manipulate generated images in a targeted way. Based on treating the model as a continuous function and by passing gradients between the image space and the prompt embedding space, we propose and analyze a new method to directly manipulate the embedding of a prompt instead of the prompt text. We then derive three practical interaction tools to support users with image generation:
\Ni
Optimization of a metric defined in the image space that measures, for example, the image style.
\Nii
Supporting a user in creative tasks by allowing them to navigate in the image space along a selection of directions of ``near'' prompt embeddings.
\Niii
Changing the embedding of the prompt to include information that a user has seen in a particular seed but has difficulty describing in the prompt.
Compared to prompt engineering, user-driven prompt embedding manipulation enables a more fine-grained, targeted control that integrates a user's intentions. Our user study shows that our methods are considered less tedious and that the resulting images are often preferred.
\end{abstract}

\section{Introduction}

Generative text-to-image models such as Stable Diffusion \cite{rombach:2022} allow their users to generate images based on a textual description called a prompt. If a generated image does not satisfy a user directly, adjusting the prompt is currently the primary \emph{targeted} way to change it to their liking. Since users have found that certain prompts are more likely to produce satisfactory images than others, several approaches to write and refine prompts have emerged. The resulting variety of prompt design patterns and best practices is collectively referred to as prompt engineering \cite{hao:2022,witteveen:2022}. As shown in the upper left of Figure~\ref{prompt-embedding-manipulation-illustration}, prompt engineering is an iterative process: In each iteration, a user assesses the image generated in the previous (or first) iteration for a given prompt, and then attempts to reformulate the prompt to achieve a desired effect. If the reformulations are successful, the user may learn how the model interprets a prompt in general, i.e., its ``prompt language.''

\bsfigure{prompt-embedding-manipulation-illustration}{Our three techniques for manipulating prompt embeddings enable a user to
\Ni
optimize an image quality metric,
\Nii
navigate the prompt embedding space towards nearby variants, and
\Niii
reconstruct a preferred image by introducing seed invariance.}

Prompt engineering has several shortcomings: The prompt language of a model is opaque to a user, and its interpretation by the model may differ arbitrarily from that of the user, due to the inherent ambiguity of natural language as well as potentially misleading correlations in the model's training data. In addition, a model may not consider the same parts of a prompt as important as the user, so that clearly phrased prompts may have little to no impact on the generated image. Moreover, certain aspects of an image are difficult to describe, such as stylistic and aesthetic aspects as well as minute details. And generative models are often used non-deterministically in that a new random seed is used to initialize inference for each new prompt submission, which can lead to unpredictable results for a prompt that worked well beforehand with a different seed. Overall, users report that they have a ``sense of direction'' during prompt engineering, but no control over the process \cite{deckers:2023a}. We attribute this to the iterative nature of prompt engineering and a fundamental mismatch between user expectations during prompt engineering and model behavior: A generative text-to-image model does not use information about a user's previous interactions in a prompt engineering session, while the user builds a mental model from their interactions to reformulate prompts. This leads users to presume predictable model behavior in situations where none can be expected. For inexperienced users, prompt engineering may therefore basically seem not much better than trial and error.

In this paper, we propose and analyze a new targeted approach to support a user in creating an image (see Figure~\ref{prompt-embedding-manipulation-illustration}). Instead of prompt engineering, we develop a technique that allows the user to directly manipulate the prompt's embedding in a meaningful way. In typical text-to-image models, a prompt is mapped into an embedding space before the corresponding image is generated. Based on our observation that small changes to the embedding of a prompt lead to small changes in the generated image, a direct manipulation of a prompt's embedding allows the continuous modification of the information originally contained in the prompt in arbitrarily fine steps. This relieves the user of verbalizing the desired changes in a generated image as well as finding a wording that the model understands, leading to a better satisfaction with each iteration. We derive three practical interaction tools that differ in the way they determine the direction in the prompt embedding space in which to modify the prompt embedding (Section~\ref{sec:3}):
\Ni
A method that optimizes a metric living in the image space that captures, for example, certain stylistic or aesthetic aspects.
\Nii
A human feedback-based method in which the user selects the direction in which to modify the prompt embedding from a list of alternatives.
\Niii
A method based on a target image generated from a prompt and a specific seed that allows the user to regenerate a similar image compared to the target image, regardless of the seed used.
These methods align with three types of creative processes, namely that of seeking to achieve a certain aesthetic, find inspiration, or reproduce existing image components. We evaluate our methods in experiments and a user study (Section~\ref{sec:4}). All the code and data for our methods are publicly available.%
\footnote{\raggedright Code: \url{https://github.com/webis-de/IJCAI-24}\\
Data: \url{https://doi.org/10.5281/zenodo.8274625}}

\section{Background and Related Work}
\label{sec:2}

To motivate the idea of prompt embedding manipulation, this section reviews the pipeline used for generative text-to-image models such as Stable Diffusion, and points to related approaches that allow the user to control the generation of the image with no or limited prompting.

\subsection{Stable Diffusion}

Stable Diffusion \cite{rombach:2022} is based on the concept of diffusion probabilistic models \cite{sohl-dickstein:2015} and implements a U-Net as an autoencoder in the denoising step to make this architecture suitable for generating images. The denoising process, which is executed to generate an image, starts with a randomly initialized latent so that a seed can be used for the generation. What makes Stable Diffusion useful as a generative text-to-image model is its conditioning mechanism. It uses a cross-attention mechanism \cite{vaswani:2017} and allows for different input modalities. For training Stable Diffusion, the LAION dataset \cite{schuhmann:2022} with text-to-image pairs was used. However, the texts (and thus also the prompts) are not used directly in the conditioning mechanism, but are first converted into embeddings using the CLIP encoder \cite{radford:2021}.

Other state-of-the-art generative text-to-image models use a similar pipeline by using either CLIP embeddings \cite{ramesh:2022,nichol:2022,podell:2023}, different encoders such as T5-XXL \cite{raffel:2019,saharia:2022,chang:2023}, or a combination of both \cite{balaji:2022}. For our experiments, we use Stable Diffusion \cite{rombach:2022}, since the model weights are publicly available. However, the model is mostly treated as a black box (except for the calculation of the gradients), so our approach can also be applied to other models.

\subsection{Interpolation of Prompt Embeddings}
\label{sec:interpolation}

\bsfigure[scale=0.92]{linear-interpolation-alternatives}{Comparison of two approaches to interpolating between two prompt embeddings. NLERP results in unevenly distributed interpolated points on the sphere. Changing its interpolation parameter results in larger adjustments to the points near the center. SLERP provides more consistent control.}

\begin{figure}[t]
\centering
\small
\providecommand{\bsscale}{}
\renewcommand{\bsscale}{0.0899}
\includegraphics[scale=\bsscale]{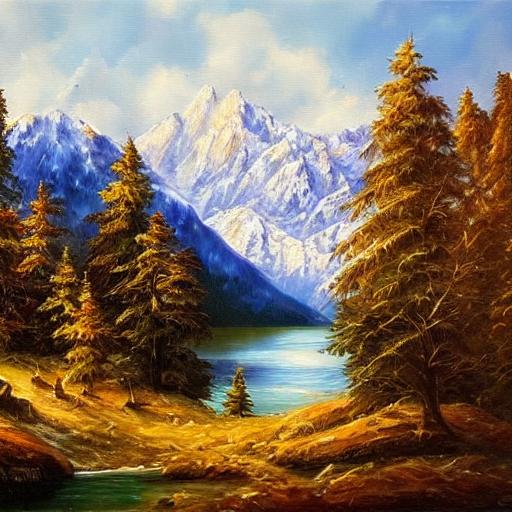}
\includegraphics[scale=\bsscale]{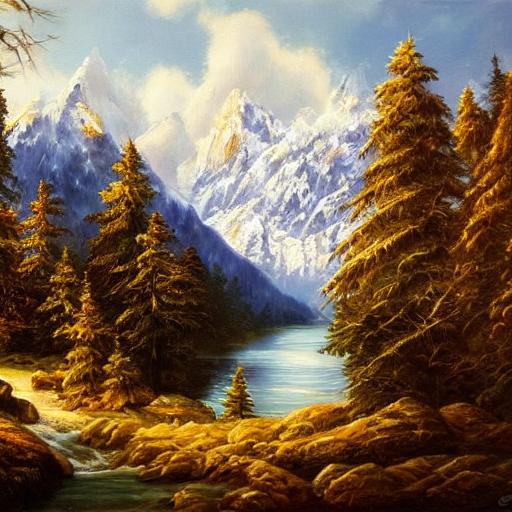}
\includegraphics[scale=\bsscale]{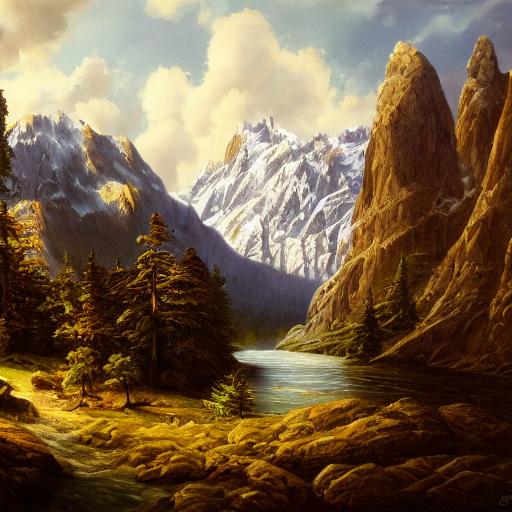}
\includegraphics[scale=\bsscale]{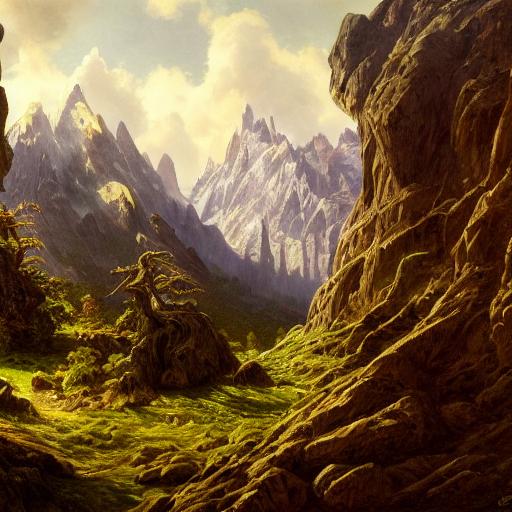}
\includegraphics[scale=\bsscale]{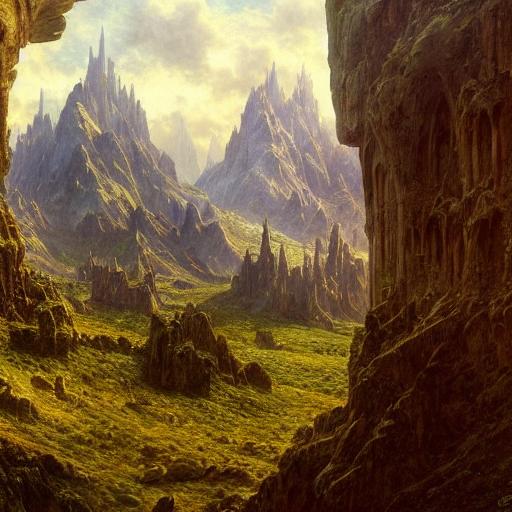}
\vspace{2ex}
\parbox[t]{0.2\columnwidth}{\centering Prompt~\ref{figure-interpolation-example-single-column}.1}%
\parbox[t]{0.6\columnwidth}{\centering \bstextrightarrow{15em}\\[-2.75ex] \fcolorbox{white}{white}{Interpolation}}%
\parbox[t]{0.2\columnwidth}{\centering Prompt~\ref{figure-interpolation-example-single-column}.2}
\vspace{-5ex}
\caption{Selected example of an interpolation between two prompts, which can be found in our published data.}
\label{figure-interpolation-example-single-column}
\end{figure}

The CLIP embeddings used by Stable Diffusion to generate images encode both the content and the style described in the prompt. Further exploring our previous idea of describing Stable Diffusion as an infinite index \cite{deckers:2023a}, we observe that the mapping from the prompt embedding space to the image space defined by Stable Diffusion is continuous in the sense that small adjustments in the prompt embedding space lead to small changes in the image space. This is true not only when considering the distance of pixel values in the images, but also for the perceived difference in content and style of the generated images. It should also be noted that small adjustments to a prompt embedding from which a high-quality image is generated will result in an image that is still of high quality.

For larger single-step adjustments, we use an interpolation between two prompt embeddings. As a consequence of the cosine similarity used to train CLIP, a linear interpolation (LERP) between the prompt embeddings is not perfect: if the norm of an embedding is not within a certain range, Stable Diffusion produces corrupted images or images with unwanted artifacts. This also means that not all values that can be specified in the same matrix format as prompt embeddings are suitable as such. Correcting the norm of linearly interpolated prompt embeddings is a practical way to avoid this problem (see NLERP in Figure~\ref{linear-interpolation-alternatives}). The use of SLERP \cite{shoemake:1985}, a spherical linear interpolation, is also possible, which is well established in connection with the interpolation of prompt embeddings \cite{han:2023}. Figure~\ref{figure-interpolation-example-single-column} shows an example of interpolation between two prompt embeddings. Since the perceived style and content are also interpolated when generating the images with Stable Diffusion, CLIP embeddings and Stable Diffusion can be considered robust in this respect.

Our proposed methods use small-step adjustments of prompt embeddings defined by gradient descent or by small steps along a SLERP interpolation, as well as larger adjustments generated by direct SLERP interpolation between the embeddings of two prompts. This allows for fine-grained and effective control when manipulating prompt embeddings. Interpolation between the initial latents (which are randomly initialized using a seed) is possible as well. 
Problems with SLERP have been observed, leading to the development of more advanced methods \cite{samuel:2023}. Nevertheless, we used SLERP because it proved to be feasible for the small-step adjustments made in our experiments.

\subsection{Related Work}
\label{sec:related-work}

For generative text-to-image models, the prompt engineering process is supported by frameworks such as the AUTOMATIC1111 Web UI,%
\footnote{\url{https://github.com/AUTOMATIC1111/stable-diffusion-webui}}
which provides useful tools for suggesting prompt modifiers or changing certain areas of the image (inpainting). One approach to providing the text-to-image model with the information that would otherwise be contained in a reformulated prompt is to allow the input of images. While inpainting is a very direct method, as it simply copies information between image areas, other methods have been introduced that allow more indirect and complex interactions with the provided images. Some of them introduce an editing tool that allows a concept to be learned from given images, which can then be referred to in user-defined prompts without having to verbally describe the learned concept in detail. This is done by finetuning the model weights \cite{ruiz:2022,han:2023} or by learning a representation of the concept in the embedding space \cite{gal:2023}. Methods such as LDEdit \cite{chandramouli:2022} calculate a latent that could be used to generate a particular image, and then generate a new image using a modified prompt. Other methods allow to modify a given generated or real image based on an original and a modified prompt, which may involve prompt engineering \cite{hertz:2023,mokady:2022,li:2023}. It has also been shown that a discrete token representation can be used to approximate a given target image \cite{wen:2023}, providing better interpretability but reducing flexibility.

To incorporate human feedback, e.g., through an aesthetics metric, it is also possible to finetune the diffusion model \cite{black:2023}. However, compared to changing the underlying prompt embedding, this is expensive and does not reflect the process of prompt engineering. Human feedback can also be used to finetune the CLIP encoder to better align the models with user preferences \cite{wu:2023}. This has a direct effect on the prompt embedding, but does not allow individual adjustments to individual prompts. Human feedback in the form of binary ratings can also be used to iteratively change the weights of the self-observation module of the U-Net, resulting in individual tuning for a given prompt \cite{ruette:2023}.

It can be helpful to provide the model with information in different input modalities. ControlNet \cite{zhang:2023} allows Stable Diffusion to be extended so that it can be finetuned to accept, e.g., segmentation maps, depth maps or human doodles. Various options for controlling diffusion models go so far as to use brain activity instead of text prompts \cite{takagi:2023}.

The optimization of prompts has also been investigated in the context of generative language models. Here, text prompts can be considered discrete, which requires special optimization methods \cite{deng:2022}. Continuous prompt embeddings were introduced, which allow training without finetuning the used model itself \cite{liu:2021,lester:2021}.

\subsection{Prompt Datasets}

Our experiments require a large number of prompts. For the evaluation and some of the illustrations, we used a subset of the prompts from DiffusionDB \cite{wang:2023}. For the user study in Section~\ref{sec:4.2}, some original prompts come from \href{https://lexica.art}{lexica.art}, a database of mature prompts from which we have removed some of the included prompt modifiers. All prompts used can be found in our published data.

\section{Optimization of Prompt Embeddings}
\label{sec:3}

This section introduces our three proposed prompt embedding manipulation methods as outlined in Figure~\ref{prompt-embedding-manipulation-illustration}, namely the directed optimization of prompt embeddings using a metric, human feedback, or a target image.

\subsection{Metric-Based Optimization}
\label{sec:3.1}

During prompt engineering, users often use prompt modifiers to achieve a certain style or aesthetic, for example, by appending phrases such as \texttt{4k high resolution award-winning image}. These modifiers tend to be highly arbitrary. Our method instead optimizes the embedding of a particular prompt with respect to a metric defined in the image space. If the user's desired style can be expressed by such a metric and its gradients can be calculated, our method can automatically improve the embedding of the prompt and provide better images to the user.

Typically, an image~$\mathcal{I}$ is generated from a prompt~$\mathcal{P}$ by embedding the prompt using a text encoder~$\psi$, and then applying the Latent Diffusion Model~(LDM):
\begin{equation}
\mathcal{I} = \mathrm{LDM}(\psi(\mathcal{P}))
\end{equation}
Given a metric~$m$ that maps images to a numeric value in a differentiable way, we use gradient descent (or gradient ascent if the metric denotes an improvement by an increasing value) to optimize the prompt embeddings with
\begin{equation}
\mathcal{C}^* = \underset{\mathcal{C}}{\mathrm{arg\,min}}\; m(\mathrm{LDM}(\mathcal{C})),
\end{equation}
where the prompt embeddings are initialized as
\begin{equation}
\mathcal{C} = \psi(\mathcal{P}).
\end{equation}
The resulting image is
\begin{equation}
\mathcal{I}^* = \mathrm{LDM}(\mathcal{C}^*).
\end{equation}

It should be noted that we do not update (i.e., finetune) the model weights during the optimization the prompt embedding~$\mathcal{C}$. Using gradient descent allows us to make relatively small changes to it, keeping most aspects of the generated image intact, while still optimizing with respect to the metric~$m$. Specifically for Stable Diffusion, it has proven helpful to optimize both the conditioning and unconditional conditioning values. During the optimization, we keep track of the seed used. However, our experiments (Section~\ref{sec:4.1}) also cover a generalization across seeds.

We implement our method for three metrics: a pair of basic metrics, blurriness and sharpness, and an advanced deep learning-based aesthetic metric. The blurriness metric is defined by converting the image to grayscale, computing the discrete Laplacian by applying a 2D~nine-point stencil via convolution, and returning the variance of the Laplacian. The sharpness metric is defined as the negative of the blurriness metric. To measure the aesthetic quality of an image in pixel space, we resort to the pre-trained LAION aesthetic predictor.%
\footnote{\url{https://laion.ai/blog/laion-aesthetics/}}$^{,}$\footnote{\url{https://github.com/christophschuhmann/improved-aesthetic-predictor}}
Its score is determined by first calculating the CLIP embedding of a given image and then feeding it into a linear model that has been trained to predict a score between~1 and~10 based on 176,000~human ratings of image aesthetics. This pipeline forms a metric that we use for describing and optimizing aesthetic quality. Note that for all three metrics, the gradients can be calculated automatically, making them suitable for the proposed method.

\subsection{Iterative Human Feedback}
\label{sec:3.2}

Generative text-to-image models are often used for creative tasks where a general theme is given, but the user does not have a specific target image in mind. Users can vary the seed to gain inspiration, but this method is quite limited and lacks control. In the context of prompt engineering, this can lead to a process of trial and error where users apply different prompt modifiers to improve their prompt locally.
Our goal is to iteratively provide inspiration to the user in the form of suggested related images based on a modified prompt embedding.

After computing the current prompt embedding as $\mathcal{C} = \psi(\mathcal{P})$ from an initial prompt $\mathcal{P}$, each step is defined as follows:
To generate choices for the user to select from, we generate prompt embeddings $\hat{\mathcal{C}}_i$ as 
\begin{equation}
    \hat{\mathcal{C}}_i = \mathrm{SLERP}(\mathcal{C}, \tilde{\mathcal{C}}_i, c_i),
\end{equation}
where the prompt embeddings $\tilde{\mathcal{C}}_i$ are generated from random prompts $\tilde{\mathcal{P}}_i$, which are mainly created by concatenating random alphanumeric characters. From a large set of such potential candidates, a subset is selected that approximates a maximum pairwise cosine distance. This creates a diverse range of prompt embedding candidates.
The interpolation parameter $c_i$ is chosen to keep $\mathcal{C} \cdot \hat{\mathcal{C}}_i$ constant and equal for each individual choice, allowing for an equal perceived distance of the choice from the current prompt embedding.

In a second step identical to the above, we modify the embedding $\hat{\mathcal{C}}_i$ towards the original prompt (also in this case modified by a randomly selected prompt modifier from a list of established modifiers). This re-introduces aesthetic quality and prevents the interactive method from diverging too much from the original meaning.

The choices given to the user are thus
\begin{equation}
    \hat{\mathcal{I}}_i = \mathrm{LDM}(\hat{\mathcal{C}}_i).
\end{equation}
The user is now able to select a choice $i$ and assign an interpolation parameter $\alpha \in \left[0,1\right]$, that is used to determine the new current prompt embedding for the next step as
\begin{equation}
    \mathring{\mathcal{C}} = \mathrm{SLERP}(\mathcal{C}, \hat{\mathcal{C}}_i, \alpha ).
\end{equation}
The new current image can now be displayed as
\begin{equation}
    \mathring{\mathcal{I}} = \mathrm{LDM}(\mathring{\mathcal{C}}).
\end{equation}

Again, this method can be considered an optimization, where each step locally optimizes the user satisfaction.
During the iterative process, we keep the used seed fixed to improve the predictability of the results.

\bsfigure[width=\columnwidth]{user-study-interface-layout}{The user interface for our iterative human feedback method. The current image is shown on the bottom left. The choices are shown at the top. The bottom right shows a t-SNE \protect\cite{maaten:2008} dimensionality reduction of the current embedding in the center and the five options scattered around.}

Figure~\ref{user-study-interface-layout} shows an implementation of the user interface for this method, allowing the user to choose between five options in each step.

\subsection{Seed-Invariant Prompt Embeddings}
\label{sec:3.3}

\begin{figure}[t]
\centering
\small
\providecommand{\bsscale}{}
\renewcommand{\bsscale}{0.0899}
\includegraphics[scale=\bsscale]{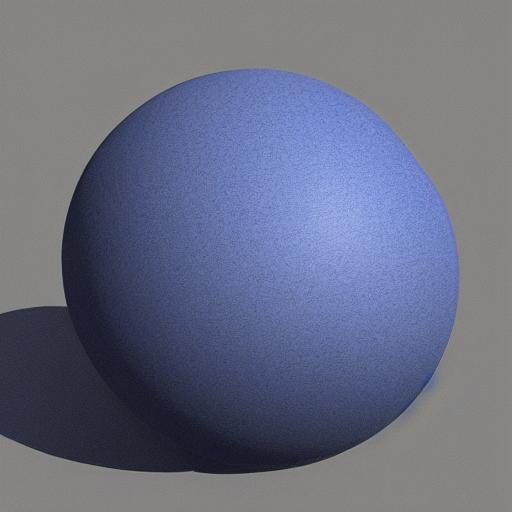}
\includegraphics[scale=\bsscale]{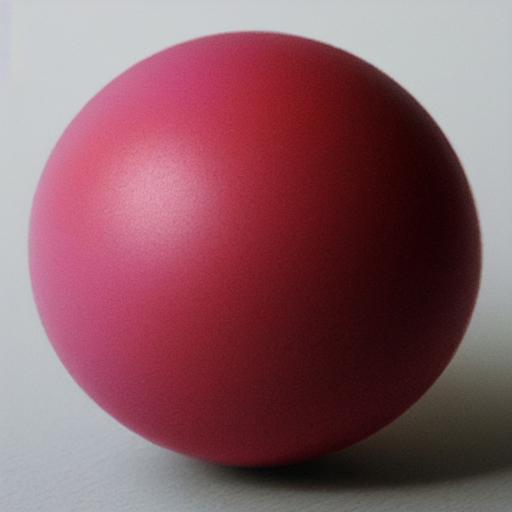}
\includegraphics[scale=\bsscale]{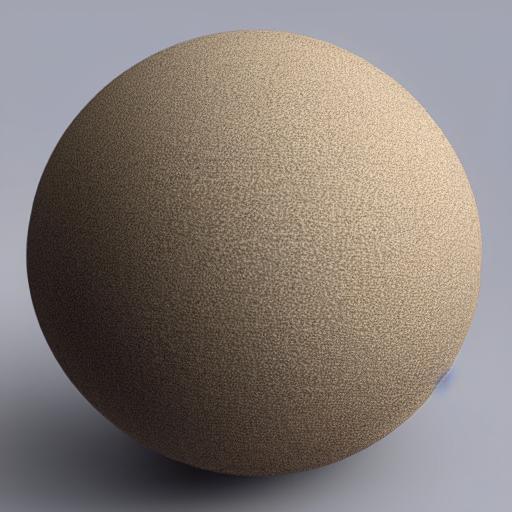}
\includegraphics[scale=\bsscale]{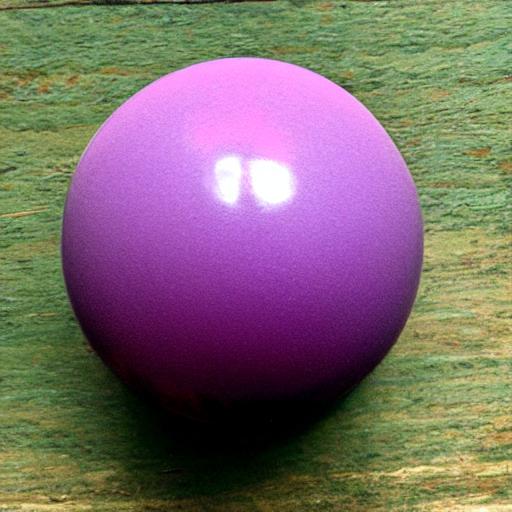}
\includegraphics[scale=\bsscale]{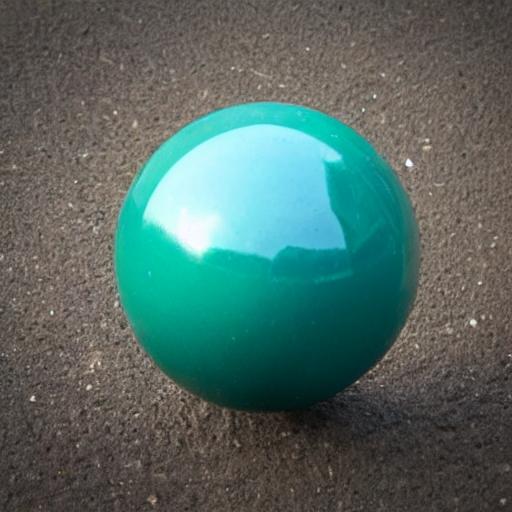}
\vspace{2ex}
\parbox[t]{0.2\columnwidth}{\centering Seed~\ref{figure-random-seed}.1}%
\parbox[t]{0.2\columnwidth}{\centering Seed~\ref{figure-random-seed}.2}%
\parbox[t]{0.2\columnwidth}{\centering Seed~\ref{figure-random-seed}.3}%
\parbox[t]{0.2\columnwidth}{\centering Seed~\ref{figure-random-seed}.4}%
\parbox[t]{0.2\columnwidth}{\centering Seed~\ref{figure-random-seed}.5}%
\vspace{-2ex}
\caption{Selected images generated with the prompt \texttt{Single Color Ball} and different random seeds.}
\label{figure-random-seed}
\end{figure}

During the process of prompt engineering, users typically try out different seeds to seek inspiration. If they discover something interesting, e.g., an object or style, the users typically try to verbalize this aspect to include it in the prompt, which can be very difficult.
As shown in Figure~\ref{figure-random-seed}, the seed can have a large effect when using certain prompts.
If the user's satisfaction depends on the seed, this may indicate that the prompt does not contain all the necessary information.
We propose an automatic method to remove the underspecification of the prompt \cite{hutchinson:2022} by modifying the prompt embeddings directly.
Unlike textual inversion \cite{gal:2023}, our method does not aim to preserve the variance induced by different seeds, e.g., to vary the perspective when showing an object. We want to describe the image induced by a single seed as specifically as possible.
This could also be useful when users aim to preserve certain aspects, e.g., how a single region or object in the image looks like, as our proposed method could be restricted to this particular aspect. This allows the rest of the image to be iteratively improved.

Given a target image $\mathcal{I}$ created using a prompt $\mathcal{P}$ and an initial latent $z$, the goal is to find a prompt embedding $\mathcal{C}^*$ such that 
\begin{equation}
    \mathrm{LDM}(\psi(\mathcal{P}), z) = \mathrm{LDM}(\mathcal{C}^*, \tilde{z})
\end{equation}
for any feasibly initial latent $\tilde{z}$.

\begin{algorithm}[t]
\caption{Seed-Invariant Prompt Embeddings}\label{alg:seed}
\begin{algorithmic}[1]
\STATE $\mathcal{I} \gets \mathrm{LDM}(\psi(\mathcal{P}), z)$
\STATE $\mathcal{C} \gets \psi(\mathcal{P})$
\FOR{$\alpha \gets \frac{1}{n}, \dots, \frac{n}{n}$}
    \STATE Sample $\tilde{z}$ as a batch of random initial latents
    \STATE $L \gets \| \mathcal{I} - \mathrm{LDM}(\mathcal{C}, \mathrm{SLERP}(z, \tilde{z}, \alpha)) \|_2^2$
    \STATE $\mathcal{C} \gets \mathcal{C} - \eta \nabla_{\mathcal{C}} L $
\ENDFOR
\RETURN $\mathcal{C}$
\end{algorithmic}
\end{algorithm}

The pseudocode in Algorithm~\ref{alg:seed} outlines the proposed method.
This algorithm uses gradient descent to optimize a loss with respect to the current prompt embedding $\mathcal{C}$, bringing its image for random seeds closer to the target image.
The random seeds are only introduced gradually using the interpolation parameter $\alpha$.
It is feasible to restrict the output of $\mathrm{LDM}(\dots)$ to only the first latents in the beginning.

\bsfigure[width=\columnwidth]{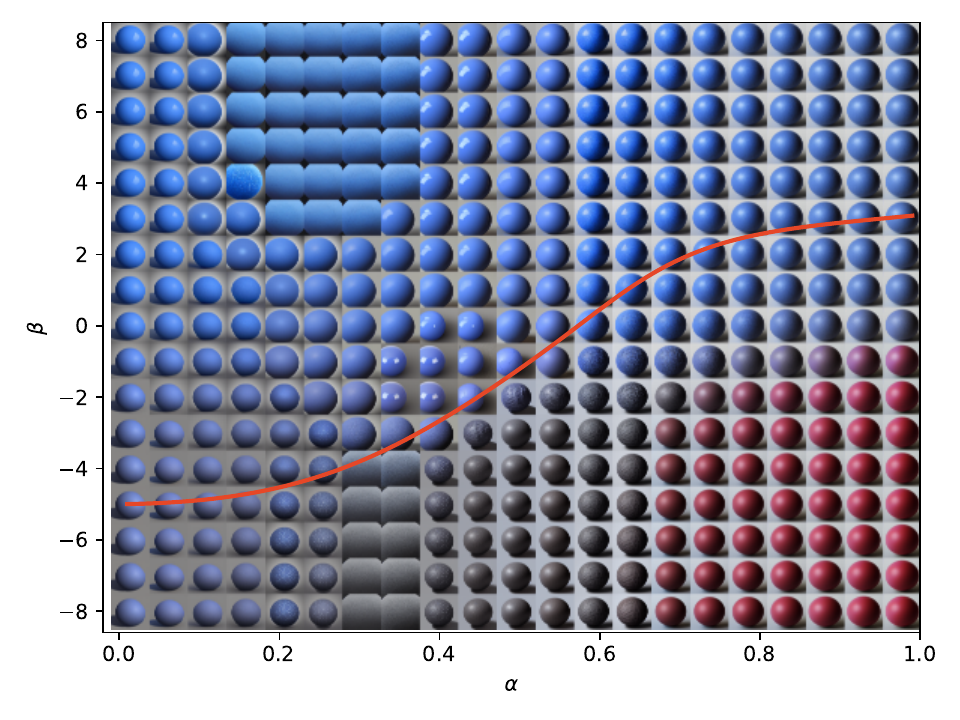}{ Traversing the prompt embedding space for a gradually modified seed. $\alpha$ denotes the SLERP interpolation parameter between two seeds Seed~\ref{figure-random-seed}.1 (left) and Seed~\ref{figure-random-seed}.2 (right). The ordinate represents the prompt embedding space with $\mathrm{sigmoid}(\beta)$ denoting the SLERP interpolation parameter between \texttt{Single Color Ball} (bottom) and \texttt{Blue Single Color Ball} (top). The orange curve denotes the learned $\beta$ for each $\alpha$ step.}

To further illustrate the proposed method, we use an oversimplified example. We reuse the images from Figure~\ref{figure-random-seed} and try to reach a prompt embedding which still shows an image like that of Seed~\ref{figure-random-seed}.1 when prompted with a different seed like Seed~\ref{figure-random-seed}.2.
We simplify the algorithm above by restricting the space for $\mathcal{C}$ to a one-dimensional interpolation between the prompt embeddings of \texttt{Single Color Ball} and \texttt{Blue Single Color Ball}. This setup is shown in Figure~\ref{traversing-simplified-prompt-space}. If our intuition about our method is correct, our $\mathcal{C}$ will move towards a prompt that encodes seed-specific information about our target image. This means that the curve in Figure~\ref{traversing-simplified-prompt-space} should move up towards a positive $\beta$ as our $\alpha$ increases. Our experiment confirms this.

\section{Experimental Results}
\label{sec:4}

Our experiments apply our methods to Stable Diffusion in different settings, measuring their success either directly, or through human feedback.%
\footnote{We used Pytorch~2.0 under Python~3.10 in a dockerized Ubuntu system on an A100~GPU. However, not the full memory of the GPU was used as Stable Diffusion is able to run with~8\,GB of VRAM. Further details can be found in our published data.}

\subsection{Metric-Based Optimization}
\label{sec:4.1}

\begin{figure}[t]
\centering
\small
\providecommand{\bsscale}{}
\renewcommand{\bsscale}{0.0899}
\providecommand{\bsexample}{}
\renewcommand{\bsexample}[1]{%
\includegraphics[scale=\bsscale]{#1/blurriness/1.jpg}
\includegraphics[scale=\bsscale]{#1/blurriness/2.jpg}
\includegraphics[scale=\bsscale]{#1/blurriness/3.jpg}
\includegraphics[scale=\bsscale]{#1/blurriness/4.jpg}
\includegraphics[scale=\bsscale]{#1/blurriness/5.jpg}
\vspace{2ex}
\parbox[t]{0.2\columnwidth}{\centering Prompt~\ref{figure-metric-example-blurriness-vs-sharpness}.1}%
\parbox[t]{0.8\columnwidth}{\centering \bstextrightarrow{20em}\\[-2.75ex] \fcolorbox{white}{white}{Metric: \ $\blacktriangle$ blurriness \ $\blacktriangledown$ sharpness}}%
\vspace{-2ex}
\includegraphics[scale=\bsscale]{#1/sharpness/1.jpg}
\includegraphics[scale=\bsscale]{#1/sharpness/2.jpg}
\includegraphics[scale=\bsscale]{#1/sharpness/3.jpg}
\includegraphics[scale=\bsscale]{#1/sharpness/4.jpg}
\includegraphics[scale=\bsscale]{#1/sharpness/5.jpg}}
\bsexample{metric-examples-blurriness-vs-sharpenss/coffee-cup-with-magma}
\caption{Selected examples of optimizing metrics blurriness (top) and sharpness (bottom).}
\label{figure-metric-example-blurriness-vs-sharpness}
\end{figure}

\begin{figure}[t]
\small
\providecommand{\bsscale}{}
\renewcommand{\bsscale}{0.0899}
\providecommand{\bsexample}{}
\renewcommand{\bsexample}[1]{
\includegraphics[scale=\bsscale]{#1/1.jpg}
\includegraphics[scale=\bsscale]{#1/2.jpg}
\includegraphics[scale=\bsscale]{#1/3.jpg}
\includegraphics[scale=\bsscale]{#1/4.jpg}
\includegraphics[scale=\bsscale]{#1/5.jpg}
}
\bsexample{metric-examples-aesthetics/a-landscape-pastel-in-the-style-of-noriyoshi}\\
\bsexample{metric-examples-aesthetics/an-armchair-made-from-an-avocado}\\
\bsexample{metric-examples-aesthetics/pencil-drawing-of-a-rubber-ducky}\\
\bsexample{metric-examples-aesthetics/realistic-spaceship-rocket-design}\\
\vspace{2ex}
\parbox[t]{0.2\columnwidth}{\centering Prompts~\ref{figure-metric-example-aesthetics}.1\mbox{-}4}%
\parbox[t]{0.8\columnwidth}{\centering \bstextrightarrow{19em}\\[-2.75ex] \fcolorbox{white}{white}{Metric: \ aesthetics}}%
\vspace{-2ex}
\caption{Selected examples of optimizing the aesthetics metric.}
\label{figure-metric-example-aesthetics}
\end{figure}

Figure~\ref{figure-metric-example-blurriness-vs-sharpness} shows the images generated from the updated prompt embeddings at selected time steps for the optimization of the blurriness and the sharpness metric for a single initial prompt. In Figure~\ref{figure-metric-example-aesthetics}, results of the optimization of the aesthetic metric are shown in a similar way. By comparing the different initial prompts it can be seen that the modified aspects of the images depend on the used prompt. Nevertheless, the results are very promising.

Note that the specific values of a metric required for an image to be perceived as optimal depend on the specific prompt used. Therefore, we propose to leave it to the user to inspect the images generated in increasing iterations so they can terminate the method. Continuing the optimization beyond this point shows that the used metrics can be prone to overfitting. For the blurriness and sharpness metrics, this results in an image with artifacts. This could indicate that the direction implied by the metric's gradient is outside of the prompt embedding space that Stable Diffusion is trained on (see Section~\ref{sec:interpolation}). The aesthetic metric does not seem to have this problem because it takes such effects into account. However, it is possible to optimize the images to the point where they no longer fit the original prompt.

\bsfigure[width=\linewidth]{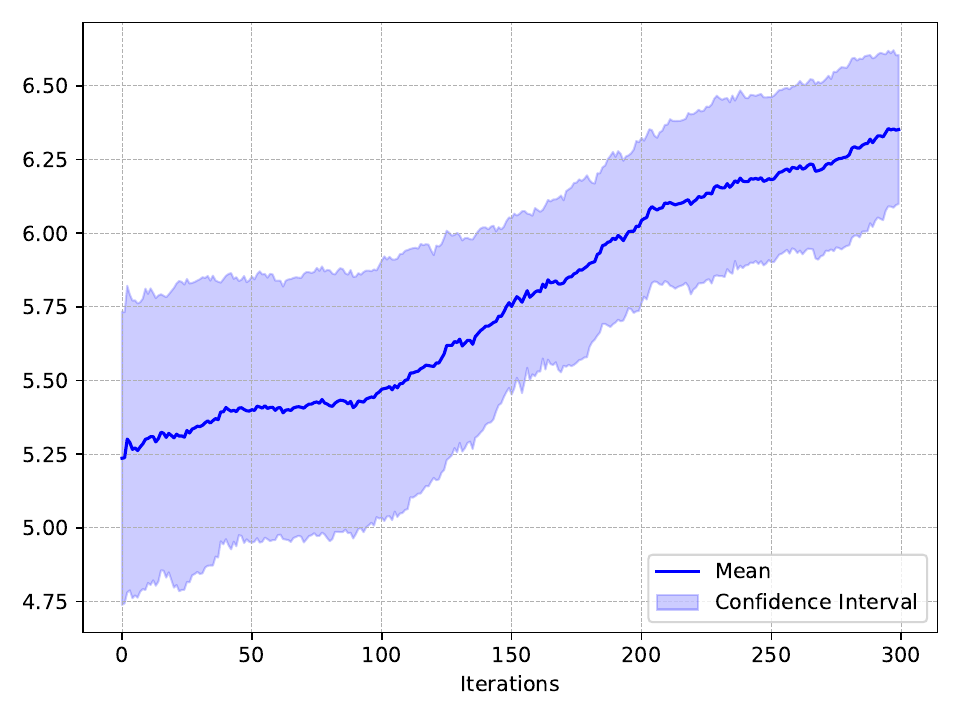}{Values of the aesthetic metric over the iteration steps of the metric-based optimization (Section~\ref{sec:3.1}) for 65 different seeds.}

When using or developing new prompt modifiers, users often want them to have the desired effect regardless of the random seed used. They sometimes need the flexibility of being able to change the seed used as a tool to adjust certain aspects of the image, such as composition, or to seek creative inspiration. Finding prompt modifiers that work independently of the seed is very helpful in this regard. We hoped to see a similar effect for our method: Despite restricting the optimization to a single seed, the modified prompt embedding should also provide an improvement regarding the metric compared to the original prompt when being applied on different seeds. To investigate this idea, we ran the optimization of the aesthetic metric for the prompt \texttt{\small highly detailed photoreal eldritch biomechanical rock monoliths, stone obelisks, aurora borealis, psychedelic} for a single seed, and stored the updated prompt embeddings for each iteration. For 65 different seeds, we now computed the values of the aesthetic metrics for these prompt embeddings. The results can be seen in Figure~\ref{metric-seed-generalization-confidence}. Not only does it show a general trend of an improving metric, it also shows a narrowing confidence interval. It can be concluded that the modified prompt embeddings are at least to some extent independent of the seed used. One could also imagine more complex methods for the optimization, which could involve multiple seeds at runtime (see Section~\ref{sec:3.3}), but the results shown are nevertheless remarkable.

\subsection{Iterative Human Feedback}
\label{sec:4.2}

\begin{figure*}
\centering
\small
\providecommand{\bsscale}{}
\renewcommand{\bsscale}{0.0899}
\providecommand{\bsexample}{}
\renewcommand{\bsexample}[1]{
\includegraphics[scale=\bsscale]{#1/ui/1.jpg}
\includegraphics[scale=\bsscale]{#1/ui/2.jpg}
\includegraphics[scale=\bsscale]{#1/ui/3.jpg}
\includegraphics[scale=\bsscale]{#1/ui/4.jpg}
\includegraphics[scale=\bsscale]{#1/ui/5.jpg}
\hspace{1em}
\includegraphics[scale=\bsscale]{#1/pe/1.jpg}
\includegraphics[scale=\bsscale]{#1/pe/2.jpg}
\includegraphics[scale=\bsscale]{#1/pe/3.jpg}
\includegraphics[scale=\bsscale]{#1/pe/4.jpg}
\includegraphics[scale=\bsscale]{#1/pe/5.jpg}
}
\bsexample{user-study/hummingbird-with-colorful-flowers}\\
\bsexample{user-study/mountain-with-a-sunset-and-a-river}\\
\bsexample{user-study/portrait-of-a-lion}\\
\vspace{2ex}
\parbox[t]{0.2\columnwidth}{\centering Prompts~\ref{figure-user-study}.1\mbox{-}3}%
\parbox[t]{0.8\columnwidth}{\centering \bstextrightarrow{15em}\\[-2.75ex] \fcolorbox{white}{white}{Our method}}%
\hspace{1em}%
\parbox[t]{0.2\columnwidth}{\centering Prompts~\ref{figure-user-study}.1\mbox{-}3}%
\parbox[t]{0.8\columnwidth}{\centering \bstextrightarrow{15em}\\[-2.75ex] \fcolorbox{white}{white}{Prompt engineering}}%
\vspace{-2ex}
\caption{Selected examples of images created in our user study using our method based on iterative human feedback and using prompt engineering. Some users achieved similar results, indicating that they were able to achieve their preferred style using our method. Other users used our method to select innovative features not seen in the prompt engineering process.}
\label{figure-user-study}
\end{figure*}

In a user study with eight participants, our method was used based on the interface shown in Figure~\ref{user-study-interface-layout} to create an image fitting a given description, following individual user preferences. For comparison with prompt engineering, we implemented a user interface similar to that of Figure~\ref{user-study-interface-layout} as a reference baseline. With each interface, the users had 20~iterations to come up with an optimal image, while half the users first used our interface, and the other half the prompt engineering interface. Throughout, our users were asked to describe their approach, and afterwards for a relative ranking between the optimal images created using both interfaces. Details can be found in our published data. Figure~\ref{figure-user-study} shows selected results.

We noticed that our method is especially helpful for creative tasks, where the user does not have a clear target image in mind. This could be discerned from the different behaviors of users who first used our method's interface versus users who first used the prompt engineering interface. The latter case can be considered a limitation of our method: User primed by prompt engineering are more dependent on being shown suggestions pointing into the direction of a desired target. Our method is also feasible for users with limited experience in prompt engineering, for whom the latter has been a rather frustrating experience. Our method was found to be less tedious, and six users preferred the image generated by our method to the one generated by prompt engineering. Contrary to the findings in Section~\ref{sec:4.1}, the prompt embeddings generated in this experiment did not generalize across the given seed, as the relative ratings seemed to differ when the seed for the optimal prompt embeddings for both methods was changed.

\subsection{Seed-Invariant Prompt Embeddings}
\label{sec:4.3}

In a less restricted experiment than the one in Section~\ref{sec:3.3}, we inspect the feasibility of our implementation for more general problems. Now, we directly optimize the high-dimensional embedding $\mathcal{C}$ without providing a low-dimensional subspace tailored for this specific experiment.

\begin{figure}[t]
\small
\providecommand{\bsscale}{}
\renewcommand{\bsscale}{0.0899}
\providecommand{\bsexample}{}
\renewcommand{\bsexample}[1]{
\includegraphics[scale=\bsscale]{#1/1.jpg}
\includegraphics[scale=\bsscale]{#1/2.jpg}
\includegraphics[scale=\bsscale]{#1/3.jpg}
\includegraphics[scale=\bsscale]{#1/4.jpg}
\includegraphics[scale=\bsscale]{#1/5.jpg}
}
\bsexample{seed-optimization/ball}\\
\bsexample{seed-optimization/cube}\\
\vspace{2ex}
\parbox[t]{0.2\columnwidth}{\centering Prompts~\ref{figure-seed-optimization}.1\mbox{-}2 \\ (Target seed)}%
\parbox[t]{0.6\columnwidth}{\centering \bstextrightarrow{10em}\\[-2.75ex] \fcolorbox{white}{white}{Optimization }\\ (Validation seed)}%
\parbox[t]{0.2\columnwidth}{\centering (Another \\ validation seed)}%
\vspace{-2ex}
\caption{Selected examples of the unguided seed-invariant prompt embedding method.}
\label{figure-seed-optimization}
\end{figure}

Figure~\ref{figure-seed-optimization} shows the first experimental results. They show that the current implementation is capable of sensing a general direction of optimization, but lacks precision, especially for complex prompts. We hope that this limitation could be overcome by borrowing implementation details from approaches like the one of \citeauthor{mokady:2022}~(\citeyear{mokady:2022}).

\section{Conclusion}

In this paper, we introduced three methods for modifying the embedding of Stable Diffusion prompts. One method optimizes a given image quality metric, another enables users to navigate the prompt embedding space, and the third allows for seed-invariant regeneration of (parts of) images. Altogether, these methods allow users to optimize their generated image directly instead of entering an iterative process of prompt engineering, avoiding trial and error. Based on our user study, we show that prompt embedding manipulation supports two types of creative tasks, one where a user looks for inspiration without having a specific target image in mind, and one where they do. Moreover, we show that manipulating prompts directly allows for optimizing image quality metrics. We believe that our methods improve the user experience when using generative text-to-image models, making them more accessible.

Future applications of our work revolve around the idea of reusing the optimized prompt embeddings. Due to their demonstrated robustness (potentially even with invariance with respect to the seeds), they can potentially be reused to improve more than one prompt. Although this would already be possible with interpolation between embeddings, different ways of integrating embedding manipulations could be investigated. Moreover, sharing optimized prompt embeddings with a community similar to, e.g., \href{https://lexica.art}{lexica.art}, appears possible. Furthermore, generalizing seed-invariant prompt embeddings towards prompt-invariance with respect to the introduced changes in manipulated prompt embeddings seems intriguing. This would lead to a single representation of, e.g., embedding manipulations toward a higher aesthetic quality, resulting in reusable embedding modifiers, which could be applied instead of the commonly used prompt modifiers.

Future research in human-computer interaction will aim to build more accessible interfaces for our methods while extending them with tools such as backtracking. Our methods can be generalized beyond Stable Diffusion, as other models have a similar architecture. The parallels to the domain of language models for text generation could potentially be used to transfer our proposed methods to this or other domains.

\newpage

\section*{Acknowledgements}

This work has been partially supported by the OpenWebSearch.eu project (funded by the~EU; GA~101070014).
\bibliographystyle{named}
\bibliography{ijcai24-manipulating-embeddings-stable-diffusion-lit}
\end{document}